\documentclass[runningheads]{llncs}
\usepackage{amsmath,amssymb,amsfonts}
\usepackage{graphicx,float}
\usepackage{ragged2e}
\usepackage{textcomp}

\usepackage[caption=false,font=normalsize,labelfont=sf,textfont=sf]{subfig}
\usepackage{float}
\usepackage{todonotes}
\usepackage[para]{footmisc}
\usepackage{booktabs}
\usepackage{xcolor,colortbl}
\usepackage{hyperref}
\usepackage{multirow}
\usepackage{cleveref}
\usepackage{rotating}
\usepackage{wrapfig}
\usepackage[noadjust]{cite}
\usepackage{paralist}
\usepackage{listings}
\usepackage{lipsum}
\usepackage{pdfpages}

\begin{document}

\title{Interpreting Black-box Machine Learning Models for high-dimensional Datasets}
\author{Anonymized for review}

\maketitle      

\begin{abstract}
     Deep neural networks~(DNNs) have been shown to outperform traditional machine learning~(ML) algorithms in a broad variety of application domains due to their effectiveness in modeling complex problems and handling high-dimensional datasets. Many real-life datasets, however, are of increasingly high-dimensionality, where a large number of features may be irrelevant for both supervised and unsupervised learning tasks. The inclusion of such features not only introduces unwanted noise impeding modeling and predicting power of a model but also increases its computational complexity. Furthermore, due to high non-linearity among a large number of features, ML models tend to be unavoidably opaque and perceived as \emph{black-box} methods. 
    In this paper, we propose an efficient method to improve the interpretability of black-box models for classification tasks in the case of high-dimensional datasets. First, we train a black-box model on a high-dimensional dataset to learn the embeddings on which the classification is performed. To decompose the inner working principles of the black-box model and to identify \emph{top-k} important features, we employ different probing and perturbing techniques. We approximate the behavior of the black-box model by training an interpretable \emph{surrogate model} on top-k feature space. Finally, we derive decision rules and local explanations from the surrogate model to explain individual decisions. Our approach outperforms state-of-the-art methods like TabNet and XGboost w.r.t metrics and explainability when tested on different datasets with varying dimensionality between 50 and 20,000.  
\end{abstract}

\section{Introduction} \label{sec:introduction}
With high availability and easy access to large-scale data, high-performance hardware accelerators, state-of-the-art machine learning~(ML) and deep neural networks~(DNNs) algorithms, predictive modeling can now be performed at scale. 
However, in the case of high-dimensional datasets including images, omics, texts, or audio, the feature space~(dimensionality) increases exponentially. In order to process such high-dimensional datasets, {singular value decomposition}~(SVD)-based approaches such as {principal component analysis}~(PCA) and {isometric feature mapping}~(Isomap) have been extensively used in the literature~\cite{fournier2019empirical}. 
These approaches were found to be moderately effective dimensionality reduction methods that 
preserve inter-point distances. However, they are fundamentally limited to linear embedding and tend to lose essential features~\cite{28}, making them less effective against curse of dimensionality~\cite{fournier2019empirical}. 
In contrast, DNNs have been shown to benefit from higher pattern recognition capabilities when it comes learning useful representation from the data. 
In particular, with multiple hidden layers and non-linear activation functions within each layer, autoencoder~(AEs) architectures can learn to encode complex and higher-order feature interactions. The resultant embeddings can capture important contextual information of underlying data~\cite{karimBIB2020}. Further, learning non-linear mappings allows transforming and mapping input feature space into a lower-dimensional latent space. Such representations can be used for both supervised and unsupervised downstream tasks~\cite{karimBIB2020}. 

Mathematical certainty means it should be possible to transparently show that there is no hidden behavior or logic influencing the behavior of a model~\cite{molnar2019interpretable}. 
However, due to high non-linearity and higher-order 
interactions among a large number of features, DNN models become \emph{black-box} because of their non-transparent structure or 
internal functioning. 
For example, latent factors learned by AEs are not easily interpretable, whilst disentangling them can provide insight into what features are captured by the representations and what attributes of the samples the tasks are based on~\cite{karimBIB2020}. Further, a serious drawback is their inability of answering questions like ``\textit{how and why inputs are ultimately mapped to certain decisions''} because the predictions made by a \emph{black-box} can neither be traced back to the input, nor it is clear why outputs are transformed in a certain way.   
Further, in sensitive areas like high-value financial trading, cybersecurity, healthcare~\cite{tjoa2020survey}, explainability and accountability are not only some desirable properties but also legal requirements~\cite{kaminski2019right} -- especially where AI may have an impact on human lives, e.g., \emph{EU GDPR} enforces that processing based on automated decision-making tools should be subject to suitable safeguards, including \textit{``right to obtain an explanation of the decision reached after such assessment and to challenge the decision''}. 

Since how a prediction is made should be as transparent as possible in a faithful and interpretable manner, model-specific and model-agnostic approaches covering local and global interpretability have emerged recently~\cite{wachter2017counterfactual}. While local explanations focus on explaining individual predictions, global explanations explain entire model behavior in the form of plots or more interpretable approximations such as decision sets or lists. 
Since interpretability comes at the cost of flexibility or efficiency, research has suggested learning a simple interpretable model to imitate the  black-box model globally and providing local explanations based on it~\cite{molnar2020interpretable}. Further, by representing the learned knowledge in a human-understandable way, an interpretable surrogate model's input-output behavior can help to better understand, explore, and validate embedded knowledge~\cite{ming2018rulematrix}, e.g., decision rules containing \emph{antecedents}~(IF) and a \emph{consequent}~(THEN) are more effective at providing intuitive explanations compared to visualization-based explanations~\cite{molnar2020interpretable}. 

Explanations that highlight the greatest difference between the object of interest and the reference object as the most user-friendly explanations~\cite{molnar2020interpretable}. Therefore, humans tend to think in a counterfactual way by asking questions like ``how would the prediction have been if input $x$ had been different?". Therefore, by using a set of rules and counterfactuals, it is possible to explain decisions directly to humans with the ability to comprehend the underlying reason so that users can focus on understanding the learned knowledge without focusing much on the underlying data representations~\cite{ming2018rulematrix}. 

However, the inclusion of a large number of irrelevant features not only introduces unwanted noise but also increase computational complexity as the data tends to become sparser. Consequently, the model tend to perform worse. 
With increased predictive modeling complexity involving hundreds of relevant features and their interactions, making a general conclusion or interpreting black-box model's outcome and their global behavior becomes increasingly difficult, whereas many existing approaches do not take into account probing or understanding the inner structure of opaque models.

Considering the drawbacks of using black-box models, we propose a novel and efficient method to improve the interpretability of black-box models for classification tasks in the case of high-dimensional datasets. 
We train a black-box model on high-dimensional datasets to learn the embeddings on which we perform the classification. 
Then, we apply probing and perturbing techniques and identify the most important top-k features. An interpretable surrogate model is then built on the simplified feature space to approximate the behavior of the black-box. \emph{decision-rules} and \emph{counterfactuals} are then extracted from the surrogate model. We hypothesize that: i) by decomposing the inner logic~(e.g., what features the model put more attention to) of black-box, its opaqueness can be reduced by outlining the most and least important features, and ii) a representative decision rule set can be generated with the surrogate model, which can be used to sufficiently explain individual decisions in a human-interpretable way. 


\section{Related Work}\label{sec:rw}
In recent years, explainable AI~(XAI) has gained a lot of attention from both academia and industries~\cite{tjoa2020survey}. Consequently, numerous interpretable ML methods and techniques have been proposed. Existing interpretable ML methods can be categorized as either model-specific or model-agnostic with a focus on local and global interpretability or either. The {local interpretable model-agnostic explanations}~(LIME)~\cite{LIME}, model understanding through subspace explanations~(MUSE)~\cite{lakkaraju2019faithful}, SHapley Additive exPlanations~\cite{SHAP}~(SHAP and its variants kernel SHAP and tree SHAP), partial dependence plots~(PDP), individual conditional expectation~(ICE), permutation feature importance~(PFI), and counterfactual explanations~(CE)~\cite{wachter2017counterfactual} are more widely used. These methods operate by approximating the outputs of an opaque ML model via tractable logic, such as game theoretic Shapley values~(SVs) or local approximation via a linear model, thereby putting emphasis on transparency and traceability of complex or black-box models~\cite{XAI_miller}.

However, these approaches do not take into account the inner structure of an opaque black-box model. Therefore, probing, perturbing, attention mechanism, sensitivity analysis~(SA), saliency maps, and gradient-based attribution methods have been proposed to understand underlying logic of complex models. Saliency map and gradient-based methods can identify relevant regions and assign importance to each input feature, e.g., pixels in an image, where first-order gradient information of a black-box model is used to produce maps indicating the relative importance of input features. Gradient-weighted class activation mapping~(Grad-CAM++)~\cite{chattopadhay2018grad} and layer-wise relevance propagation~(LRP)~\cite{LRP1} are examples\footnote{LRP demo by Fraunhofer HHI: \url{https://lrpserver.hhi.fraunhofer.de/}} of this category that highlight relevant parts of inputs, e.g., images to a DNN which caused the decision can be highlighted. 

Attention mechanisms are proposed to improve accuracy in a variety of supervised and language modeling tasks, as they can detect larger subsets of features. Self-attention network~(SAN)~\cite{vskrlj2020feature} is proposed to identify important features from tabular data. TabNet~\cite{arik2021tabnet} is proposed, which uses a sequential attention mechanism to choose a subset of semantically meaningful features to process at each decision step. It also visualizes the importance of features and how they are combined to quantify the contribution of each feature to the model enabling local and global interpretability. SAN is found effective on datasets having a large number of features, while its performance degrades in the case of smaller datasets, indicating that having not enough data can distill the relevant parts of the feature space~\cite{vskrlj2020feature}. Since interpretability comes at cost of accuracy vs. complexity trade-off, research has suggested learning a simple interpretable model to imitate a complex model~\cite{molnar2020interpretable}. In particular, model interpretation strategies are used, which involves training an inherently interpretable \emph{surrogate} model to learn a locally faithful approximation of the black-box model~\cite{molnar2020interpretable}. Since an explanation relates feature values of a sample to its prediction, rule-based explanations are easier to understand for humans~\cite{ming2018rulematrix}. Anchor~\cite{ribeiro2018anchors} is a rule-based method that extends LIME, which provides explanations in the form of decision rules. Anchor computes rules by incrementally adding equality conditions in the antecedents, while an estimate of the rule precision is above a  threshold~\cite{guidotti2018local}. 

However, a critical drawback of rule-based explanations could arise due to overlapping and contradictory rules. Sequential covering~(SC) and Bayesian rule lists~(BRL) are proposed to deal with these issues. SC iteratively learns a single rule covering the entire training data rule-by-rule, by removing the data points that are already covered by new rules~\cite{molnar2020interpretable}, while SBRL combines pre-mined frequent patterns into a decision list using Bayesian statistics~\cite{molnar2020interpretable}. \emph{Local rule-based explanations}~(LORE)~\cite{guidotti2018local} is proposed to overcome these issues. LORE learns an interpretable model of a neighborhood generated based on genetic algorithms. Then it derives local explanations from the interpretable model. Finally, it provides explanations in the form of a \emph{decision rule}, and a set of \emph{counterfactual rules} - that signifies making what feature values may lead to a different outcome. 
While LIME indicates where to look for a decision from categorical or continuous feature values, \emph{LORE}'s counterfactual rules provide high-level and minimal-change contexts for reversing the outcome of a model. 

\section{Methods}\label{sec:methods}
The problem of classifying a set of instances is grouping them into specific classes. 
Assuming the dataset is very high-dimensional having a large feature space, we classify them on their embedding space instead original feature space. A black-box model is trained on a high-dimensional dataset to learn an embedding space used for classification. To decompose its inner working principles, we apply probing and perturbing techniques and identify \emph{top-k} features. An interpretable surrogate model is then built on top-k feature space to approximate the predictions of the black-box. Finally, \emph{decision-rules} and counterfactuals are generated from the surrogate model for each instance to explain individual decisions. 

\begin{figure*}[ht!]
	\centering
		\includegraphics[width=\textwidth]{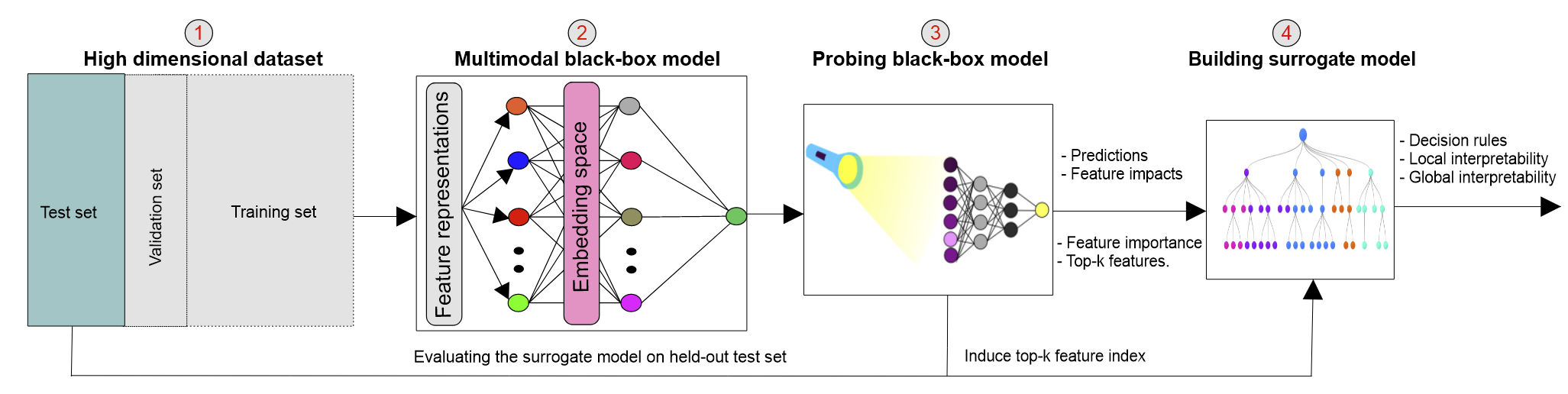}
	    \caption{Workflow of our proposed approach for interpreting black-box models, where the gray circles with red numbers represent different steps of the process.} 
	    \label{fig:workflow}
\end{figure*}

\subsection{Building black-box models} 
\Cref{fig:workflow} shows the workflow of our proposed approach for interpreting black-box models. 
Input $X$ is first fed into a DNN to generate latent representations, thereby embedding the feature space into a lower dimensional latent space, s.t. $X$ is transformed with a nonlinear mapping $f_{\Theta}: X \rightarrow Z$, where $Z \in \mathbb{R}^{K}$ are learned embeddings, where $K \ll F$. A fully-connected softmax layer is then added on top of DNN by forming a black-box classifier $f_b$. 

To parameterize $f_b$, we train a convolutional autoencoder~(CAE). Due to function approximation properties and feature learning capabilities in CAE, deep and quality features can be easily extracted~\cite{xie2016unsupervised,karimBIB2020}. Further, since weights are shared among layers, CAEs preserve locality and are able to better reduce the number of parameters in comparison to other AEs. A convolutional layer calculates the feature maps~(FMs) on which max-pooling is performed to downsample them by taking the maximum value in each non-overlapping sub-region, thereby mapping and transforming $X$ into a lower-dimensional embedding space $Z$ as follows~\cite{karimBIB2020}: 

\begin{equation}
    Z = g_\phi \left({X}\right)=\sigma\left(W \oslash X+b\right),
    \label{eq:fcuk_1}
\end{equation}

where encoder $g(.)$ is a sigmoid function parameterized by $\phi \in \Theta$ that include a weight matrix $W \in \mathbb{R}^{p \times q}$ and a bias vector $b \in \mathbb{R}^{q}$ in which $p$ and $q$ are numbers of input and hidden units, $\oslash$ is the convolutional operation, $Z$ are the latent variables, and $\sigma$ is the exponential linear unit activation function. 
The decoder function $h(.)$ reconstructs the original input $X$ from latent representation $Z$ by applying unpooling and deconvolution operations, such that $Z$ is mapped back to a reconstructed version $X^\prime \approx {X}$ as follows~\cite{karimBIB2020}: 

\begin{equation}
    X^\prime=h_{\Theta}\left(Z\right)=h_{\Theta}\left(g_{\phi}({X})\right) \label{eq:mcae_eq_1},
\end{equation}

where $h(.)$ is parameterized by $(\theta,\phi) \in \Theta$ that are jointly learned to generate $X^\prime$. This is analogous to learning an identity function, i.e., $X^\prime \approx h_{\theta}\left(g_{\phi}({X})\right)$. The mean squared error~(MSE) is used to measure the reconstruction loss $L_r$:  

\begin{equation}
    L_{\mathrm{r}}(\theta, \phi)=\frac{1}{N} \sum_{i=1}^{N}\left({X}-X^\prime \right)^{2} +\lambda\left\|W\right\|_{2}^{2},
\end{equation} 

where $\lambda$ is the activity regularizer and $W$ is a vector containing network weights. Therefore, $h_{\theta}\left(g_{\phi}({X})\right)$ is equivalent to $\Psi \left(W^\prime * Z + b^\prime \right)$~\cite{karimBIB2020}: 

\begin{equation}
    {X}^\prime=\Psi \left(W^\prime \odot Z + b^\prime \right),
\end{equation}

where $\odot$ is the transposed convolution operation, $W^\prime$ is decoder's weights, $b^\prime$ is bias vectors, and $\Psi$ is the sigmoid activation function. The unpooling is performed with switch variables~\cite{zeiler2011adaptive} such that it remembers the positions of the maximum values during the max-pooling operation. Within each neighborhood of $Z$, both value and location of maximum elements are recorded: pooled maps store values, while switches record the locations. 

The latent vector $Z$ is feed into a fully-connected softmax layer for the classification, which maps a latent point $z_i$ into an output $f_b(z_i) \mapsto \hat y_i$ in the embedding space $Z$ by optimizing categorical cross-entropy~(CE) loss~(binary CE in the case of binary classification) during back-propagation. 
The reconstruction loss and CE loss of CAE are combined and optimized jointly, during the backpropagation~\cite{karimBIB2020}:

\begin{align}
    L_\mathit{cae}=\sum_{i=1}^{n} \alpha_{r} {L_r}+\alpha_{ce} {L}_\mathit{ce},
\end{align}

where $\alpha_{r}$ and $\alpha_{ce}$ are the regularization weights assigned to reconstruction and CE loss functions, respectively.  
 
 \begin{figure*}[ht!]
	\centering
	\includegraphics[width=0.6\textwidth]{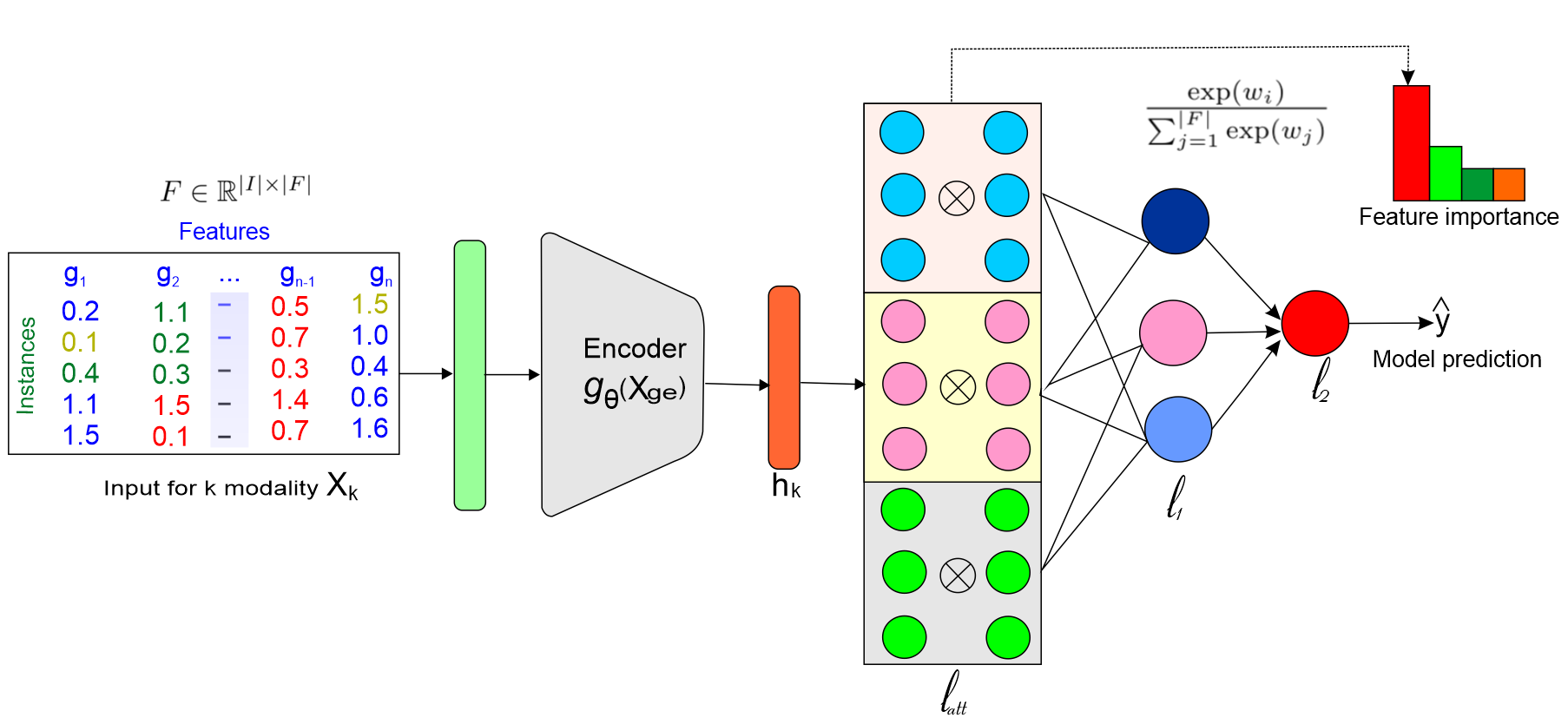}	
	\caption{Schematic representation of self-attention networks, conceptually recreated based on original SAN by Vskrlj et al.~\cite{vskrlj2020feature}.}
	\label{fig:k_attention}
\end{figure*}

\subsection{Interpreting black-box models}
We interpret the black-box model by applying probing, perturbing and model surrogation techniques.  

\subsubsection{Probing with attention mechanism}
the black-box model is probed using attention mechanism. The SAN architecture shown in \cref{fig:k_attention}, enables self-attention at the feature level. The attention layer can be represented as follows~\cite{vskrlj2020feature}: 

\begin{equation}
    l_{2}=\sigma\left(W_{2} \cdot\left(\alpha \left(W_{|F|} \cdot \Omega(X)+{b}_{l_{1}}\right)\right)+{b}_{l_{2}}\right),
\end{equation}

where $\alpha$ is an activation function, $b_{li}$ represents layerwise bias, and $\Omega$ represents the  first network layer that maintains the connection with the input features $X$~\cite{vskrlj2020feature}: 

\begin{equation}
   \Omega(X)=\frac{1}{k} \bigoplus_{k}\left[X \otimes \operatorname{softmax}\left(W_{l_{\mathrm{att}}}^{k} X+{b}_{l_{\mathrm{att}}}^{k}\right)\right],
\end{equation}

where $X$ are first used as input to a softmax-activated layer by setting: i) its number of neurons equal to $|F|$ and ii) number of attention heads representing relations between input features to $k$, and $W_{l_{\mathrm{att}}}^{k}$ represents a set of weights in individual attention heads. The softmax function applied to ${i}$-th element of a weight vector $v$ is defined as follows~\cite{vskrlj2020feature}:

\begin{equation}
    \operatorname{softmax}\left({v}_{i}\right)=\frac{\exp \left({v}_{i}\right)}{\sum_{i=1}^{|F|} \exp \left({v}_{i}\right),}
\end{equation}

where $W_{l_{\mathrm{att}}}^{k} \in \mathbb{R}^{|F| \times|F|}$, $v \in \mathbb{R}^{|F|}$, and $l_{\text {att}}$ represents the attention layer in which element-wise product with $X$ is computed in the forward pass to predict labels $\hat{y}$ of individual instances in which two consecutive dense layers $l_{1}$ and $l_{2}$ contribute to predictions,  
$\otimes$ and $\oplus$ are Hadamard product- and summation across $k$ heads. 
As $\Omega$ maintains a bijection between the set of features and attention head's weights, weights in the $|F| \times|F|$ matrix represent the relations between features. 

We hypothesize that a global attention weight vector can be generated by applying attention mechanism to encoder's bottleneck layer. This vector is then used to compute feature attributions. Unlike SAN, we apply attention to embedding space~(encoder's deepest convolutional layer), which maintains the connection between latent features $Z$ as follows: 

\begin{equation}
    \Omega(Z)=\frac{1}{k} \bigoplus_{k}\left[Z \otimes \operatorname{softmax}\left(W_{l_{\mathrm{att}}}^{k} Z+{b}_{l_{\mathrm{att}}}^{k}\right)\right].
\end{equation}

That is, embedding space $Z$ is used as the input to the softmax layer in which number of neurons is equal to dimension of embedding space. 
The softmax function applied to $i$-th element of weight vector $v_z$ as~\cite{vskrlj2020feature}:

\begin{equation}
    \operatorname{softmax}\left({v}_{{z}_{i}}\right)=\frac{\exp \left({v}_{{z}_{i}}\right)}{\sum_{i=1}^{|Z|} \exp \left({v}_{{z}_{i}}\right)};{v}_{{z}_{i}} \in \mathbb{R}^{|Z|}
\end{equation}

Once the training is finished, weights of the attention layer’s are activated using softmax activation as~\cite{vaswani2017attention}:

\begin{equation}
    R_{l}=\frac{1}{k} \bigoplus_{k}\left[\operatorname{softmax}\left(\operatorname{diag}\left(W_{l_{\mathrm{att}}}^{k}\right)\right)\right], 
\end{equation}

where $W_{l_{\mathrm{att}}}^{k} \in \mathbb{R}^{|Z| \times|Z|}$. As the surrogate model is used to provide local explanations, top-k features are extracted as diagonal of $W_{l_{\mathrm{att}}}^{k}$ and ranked w.r.t. their respective weights. 

\subsubsection{Perturbing with sensitivity analysis}
we validate globally important features through SA: we change a feature value by keeping other features unchanged. If any change in its value significantly impacts the prediction, the feature is considered to have~(statistically) high impact on the prediction. We create a new set $\hat X^{*}$ by applying $w$-perturbation over feature $a_i$ and measure its sensitivity at the global level. To measure the change in the predictions, we observe the MSE between actual and predicted labels and compare the probability distributions over the classes\footnote{Two most probable classes in a multiclass settings.}. The sensitivity $S$ of a feature $a_i$ is the difference between the MSE at original feature space $X$ and sampled $\hat X^{*}$. However, since SA requires a large number of calculations~(i.e., $N \times M$; $N$ and $M$ are the number of instances and features) across predictions, we make minimal changes by taking into consideration of only the top-k features, thereby reducing the computational complexity. 

\subsubsection{Model surrogation}
Model surrogation is about approximating the predictive power of a black-box model. Since most important features are already identified by the black-box $f_b$, we hypothesize that training a surrogate model $f$ on top-k feature space would be reasonable. We train $f$ on sampled data $X^{*}$\footnote{A simplified version of the dataset containing important features only.} and ground truths ${Y}$. Further, since any interpretable model can be used for function $g$~\cite{molnar2020interpretable}, we train decision tree~(DT) classifier. Besides, we train random forest~(RF) and extremely randomized trees~(ERT) classifiers\footnote{Eventhough RF and ERT are very complex tree ensembles and known to be black-boxes, DTs can be extracted from both classifiers. Therefore, we argue that the best DTs can be subsequently used for computing the FI.}. DT iteratively splits $X^{*}$ into multiple subsets w.r.t to threshold values of features at each node until each subset contains instances from one class only, where the relationship between prediction $\hat{y}^{*}_{i}$ and feature $a^{*}_i$ can be defined as~\cite{di2019surrogate}: 

\begin{align}
    \hat{y}^{*}_{i}=f\left(a^*_{i}\right)=\sum_{j=1}^{N} c_{j} I\left\{a^*_{i} \in R_{j}\right\},
\end{align}

where each sample $x^{*}$ reaches exactly one leaf node, $R_{j}$ is the subset of the data representing the combination of rules at each internal node, and $I\{.\}$ is an identity function~\cite{di2019surrogate}. The mean importance of a feature $a_i$ is computed by going through all the splits for which $a_i$ was used and adding up how much it has improved the prediction in a child node~(than its parent node) w.r.t decreased measure of Gini impurity $I_{GQ}= \sum_{k=1}^{N} p_{k} \cdot\left(1-p_{k}\right)$, where $p_{k}$ is the proportion of instances having label $y^*_{k}$ in a node $Q$. The sum of all the values of importance is scaled to 1. 
The RF and ERT are trained to assemble randomized decisions to generate the final decision.

\subsection{Generating feature importance and decision rules}
We assume that the black-box $f_b$ has sufficient knowledge and the \emph{surrogate} model $f$ has learned the mapping $Y^{*}=f(X^{*})$. Thus, it is reasonable to assume that surrogate model $f$ is able to mimic the model $f_b$: we compute the permutation feature importance~(PFI) for $f$ to provide a view on global feature importance~(GFI). However, since PFI does not necessarily reflect the intrinsic predictive value of a feature~\cite{lundberg2017consistent}, features having lower importance for an under-/overfitted model may turn out to be important for a better-fitted model. Therefore, we employ SHAP to generate more consistent explanations. The SHAP importance for $a_i \in x$ is computed using model $f$ by comparing what it predicts with and without the feature itself for all possible combinations of M-1 features~(i.e., except for $a_i$) in the dataset w.r.t SV $\phi_i$~\cite{lundberg2017consistent}. 

Since the order in which features are observed by a model matters, SVs explain the output of a function as the sum of effects $\phi_i$ of each feature being observed into a conditional expectation. If $a_i$ has zero effect on the prediction, it is expected to produce an SV of 0. If two features contribute equally, their SVs would be the same~\cite{SHAP}. To compute the GFI, absolute SVs per feature across all the instances are averaged~\cite{lundberg2017consistent}. Further, to make the GFI consistent, we create apply a stacking ensemble of SHAP values by averaging their marginal outputs from the DT, ERT, and RF models. 

\begin{figure*}[ht!]
	\centering
		\includegraphics[width=0.5\textwidth]{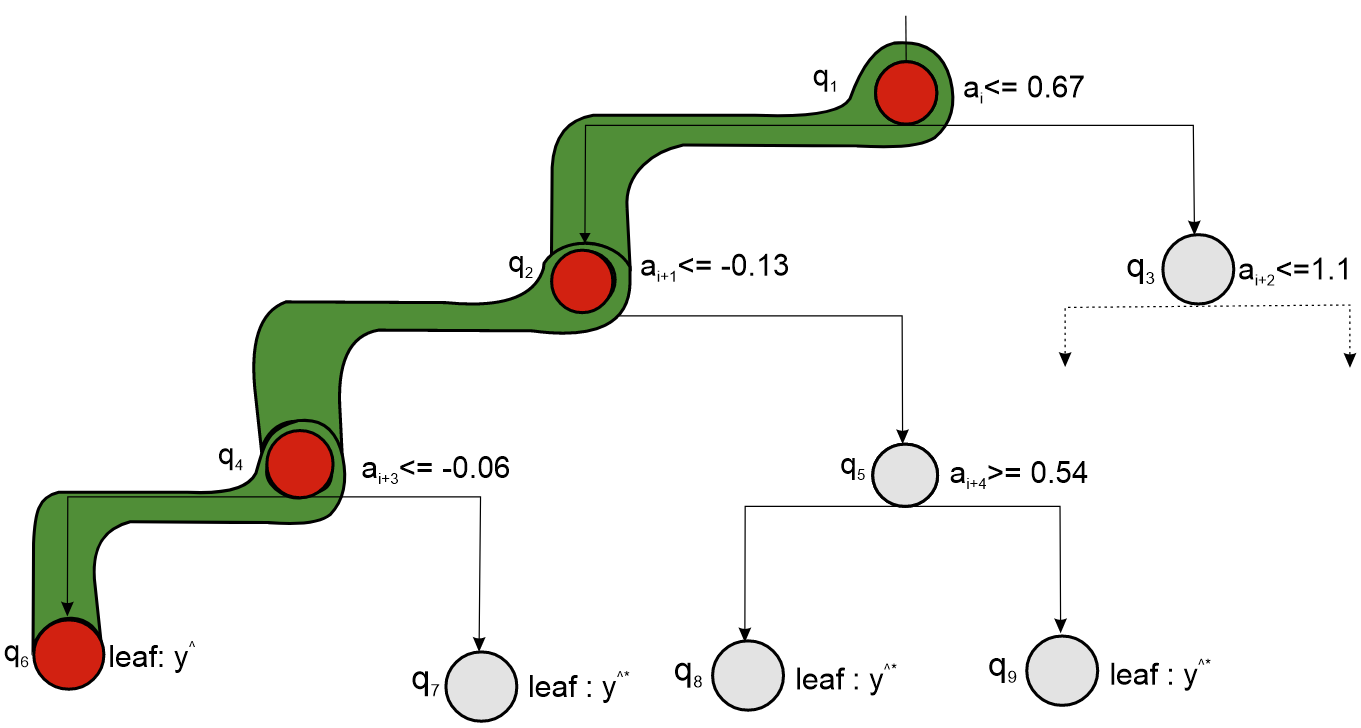}
	    \caption{Extraction of decision rules: starting at the root and satisfying the split condition of each decision node, we traverse until a leaf node is reached.} 
	    \label{fig:example_DT_path}
\end{figure*}

We extract rule list $R$ representing important rules for the surrogate ${f}$. As the decision rules can naturally be derived from a root-leaf path in a DT to predict the class for an unseen instance~\cite{guidotti2018local}, we follow the path of that instance down to the tree~(as \cref{fig:example_DT_path}). Starting at the root and satisfying the split condition of each decision node, we traverse until a leaf node is reached. Unlike in a DT, a decision can be reached through multiple rules with excessive lengths. Considering the number of predictors, we apply the representation of the ordered list of inclusive rules based on SBRL in order to solve the issue of overlapping rules. Since the feature space is huge, textual representations would reduce human-interpretability -- especially when the rule list length is large. 

Further, rules having low confidence are insignificant in discriminating classes and may not be useful in explaining individual decisions. Therefore, we filter out rules that do not meet coverage, support, and confidence thresholds. Besides, we restrict the antecedents to be the conjunction of clauses~(i.e., a condition on feature $a_i$). The output of each rule is a probability distribution over classes\footnote{Probability of an instance satisfying the antecedent belonging to a class.}. Using SBRL, pre-mined frequent patterns are combine into a decision list, where support of a pattern is computed as~\cite{molnar2020interpretable}:

\begin{align}
    \begin{array}{cl}
        \text {Support}\left(a_i \right)=\frac{1}{N} \sum_{i=1}^{N} I\left(a_{i}\right),
    \end{array}
    \label{eq:support_eq}
\end{align}

where $I \in[0,1]$, $N$ is the number of data points and $I$ is the indicator function that returns $1$ if the feature $a_{i}$ has level $i$, otherwise $0$. 
Finally, the faithfulness of rules is computed w.r.t \emph{coverage} that maximizes the fidelity of the rule list~\cite{ming2018rulematrix}.  

\begin{align}
    \text {fidelity}(R)_{{X^{*}}}=\frac{1}{|{X^{*}}|} \sum_{{x^{*}} \in {X}}[f({x^{*}})=R({x^{*}})],
    \label{eq:fidelity}
\end{align}

where $[f({x^{*}})=R(x^{*})]$ returns 1 if $f(x^{*})=R(x^{*})$, 0 otherwise. Similar to~\cite{grath2018interpretable}, we generate counterfactuals by calculating the smallest possible changes $(\Delta x)$ to input $x$ such that the outcome flips from original prediction $y$ to $y^{\prime}$. 



\section{Experiments}\label{sec:exp}
We evaluate our approach on a number of datasets, covering classification tasks. However, our approach is dataset-agnostic and can be applied to any tabular dataset. We implemented our methods in Python using \emph{scikit-learn}, \emph{Keras}, and \emph{PyTorch}. 
\subsection{Experiment settings}
To provide a fair comparison, we train TabNet and XGBoost classifiers on the same data as they are effective for tabular datasets. Besides, we train MLP on PCA projection space. Isomap outperformed for a number of tasks even though it has the computational complexity of $O(n^3)$~\cite{fournier2019empirical}. We train XGBoost on Isomap projection space. Adam optimizer and Glorot uniform initializer are used for training TabNet, MLP, and $SAN_{CAE}$ models whose hyperparameters were selected through 5-fold cross-validation and random search. 

Each model is trained on 80\% of data on Nvidia Quadro RTX 6000 GPU, followed by evaluating them on 20\% held-out set. We provide qualitative and quantitative evaluations of each model, covering both local and global explanations. We report precision, recall, F1-score, and \emph{Matthews correlation coefficient}~({MCC}) scores. We report $R^2$ to quantitatively assess the predictive power of the surrogate models and support and fidelity to assess the quality of generated rules. 
Further, to measure how well $f$ has replicated model $f_b$, R-squared measure~($R^2$) is calculated as an indicator for goodness-of-fit, which is conceptualized as the percentage of the variance of $f$ captured by the surrogate~\cite{molnar2020interpretable}: 

\begin{equation}
    R^{2}=1-\frac{SSE^{*}}{SSE} = 1 - \frac{{\sum_{i=1}^{N}}\left(\hat{y}^{*}_{i}-\hat{y}_{i}\right)^{2}}{\sum_{i=1}^{N}\left(\hat{y}_{i}-{\hat{y}}\right)^{2}},
    \label{eq:r_squared_1}
\end{equation}

where $\hat{y}^{*}_{i}$ is the prediction for $f$, $\hat{y}_{i}$ is the prediction for $f_b$ for $X^{*}$, and $SSE$ and $SSE^{*}$ are the sum of squares errors for $f$ and $f_b$, respectively~\cite{molnar2020interpretable}. Based on $R^2$, we decide if $f$ can be used for real-life use case instead of $f_b$~\cite{molnar2020interpretable}:

\begin{itemize}
    \item If $R^2$ is close to $1$~(low $\mathrm{SSE}$), the surrogate model approximates the behavior of the black-box model very well. Hence, $f_b$ can be used instead of $f$. 
    \item If $R^2$ is close to 0~(high $\mathrm{SSE}$), the surrogate fails to approximate the black-box, hence cannot replace $f_b$. 
\end{itemize}

\subsection{Datasets}
We experimented on four datasets: i) gene expression~(GE) from the \emph{Pan Cancer Atlas} project has GE values of 20,531 features, covering 33 prevalent tumor types. This dataset is  widely used as a comprehensive source of omics data for cancer type prediction and tumor-specific biomarker discovery~\cite{hoadley2018cell}. ii) indoor localization dataset~(UJIndoorLoc)~\cite{indoor_localization_dataset} having 523 variables. This multi-building indoor localization dataset is used for indoor positioning systems based on WLAN/WiFi fingerprints, iii) \emph{health advice} dataset~\cite{health_chatbot} having 123 variables. The dataset was used to provide advice based on symptoms via a healthcare chatbot, and ii) \emph{forest cover type} dataset~\cite{forest_type_dataset} having 54 variables. This dataset is used for predicting forest cover type from cartographic variables.

\subsection{Model performance analyses}
\begin{figure*}[ht!]
  \subfloat[]{
    \includegraphics[clip,width=0.25\textwidth]{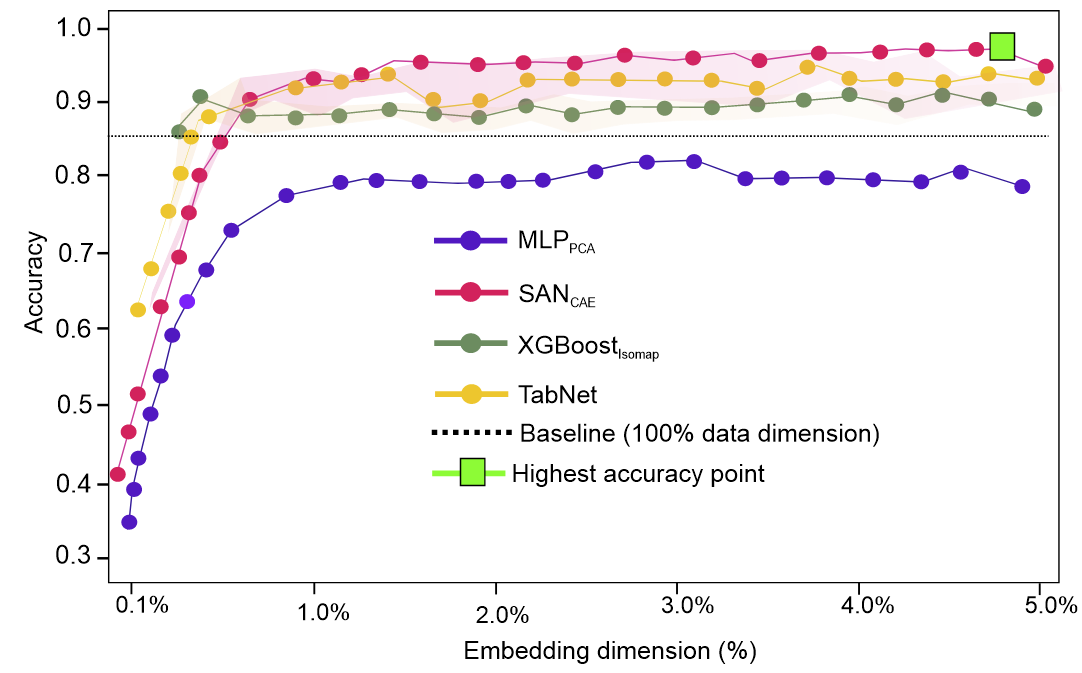}%
      \label{fig:acc_vs_dim_ge}
  }
  \subfloat[]{
    \includegraphics[clip,width=0.25\textwidth]{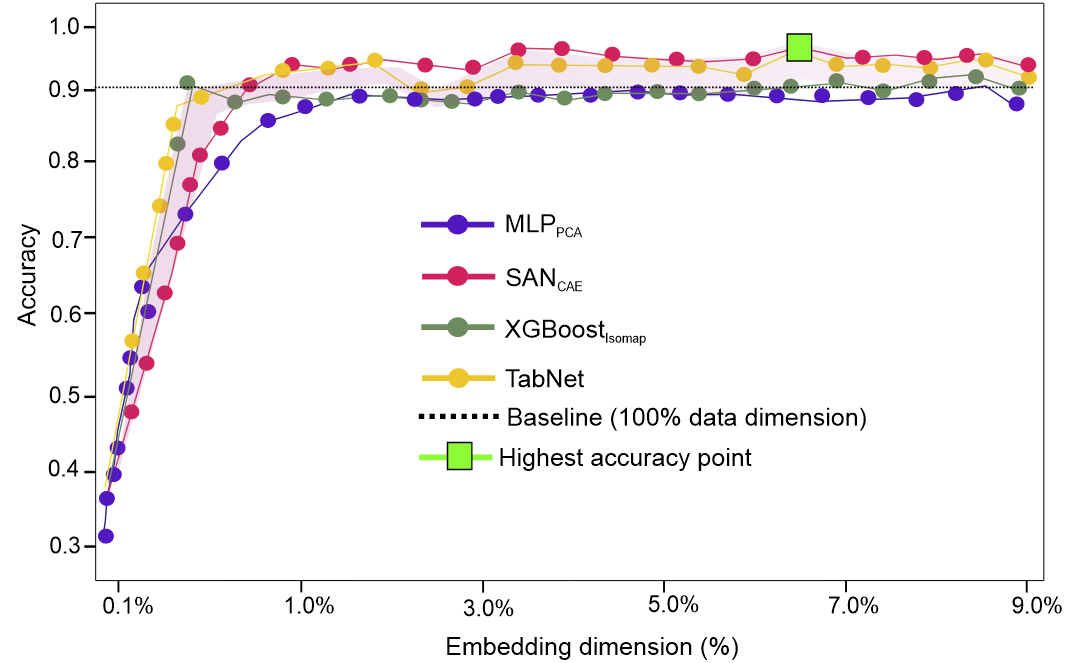}%
      \label{fig:acc_vs_dim_il}
  }
    \subfloat[]{
    \includegraphics[clip,width=0.25\textwidth]{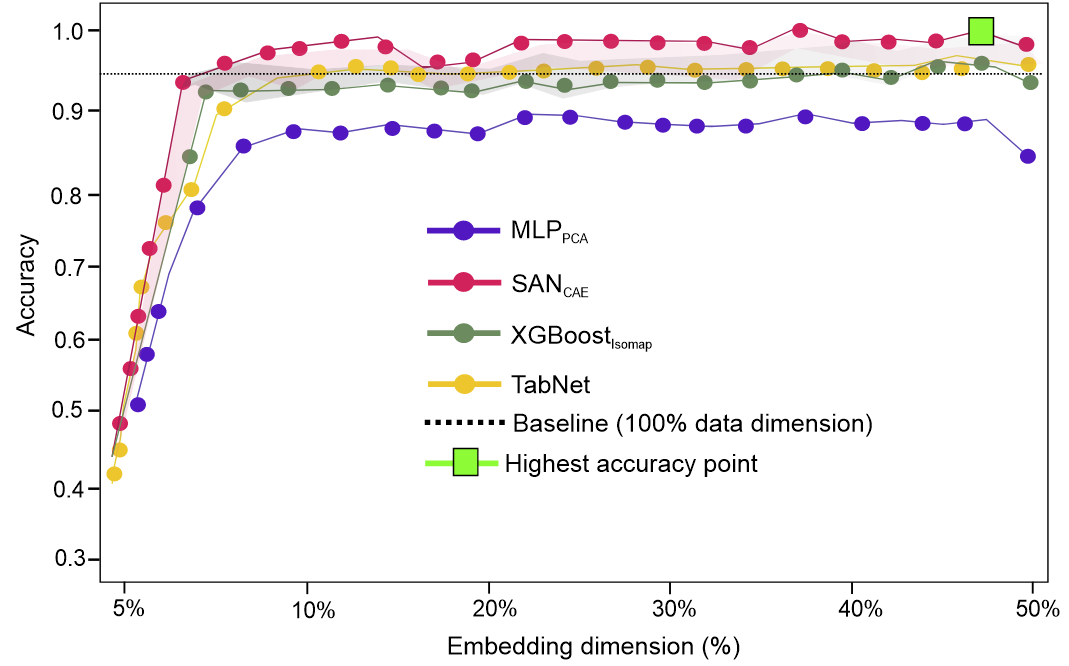}%
    \label{fig:acc_vs_dim_sp}
  }
  \subfloat[]{
    \includegraphics[clip,width=0.25\textwidth]{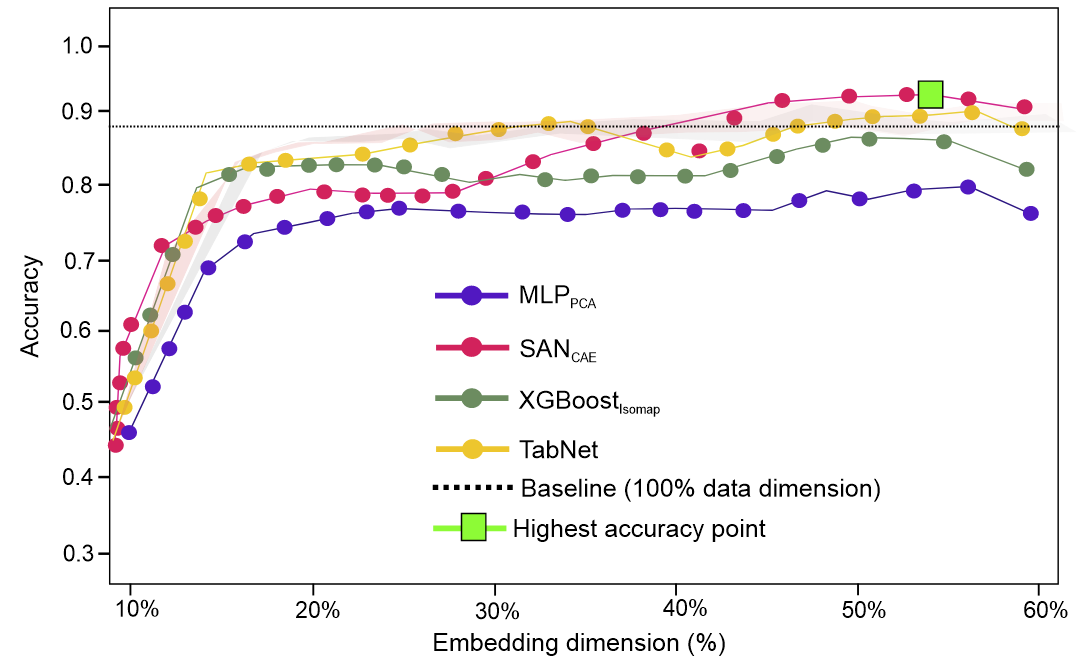}%
      \label{fig:acc_vs_dim_fct}
  }
  \caption{Mean accuracy w.r.t the relative dimension of latent space across datasets: a) gene expression, b) UJIndoorLoc, c) Health advice, d) Forest cover type. The shade indicates standard deviation. The baseline is obtained by training the $TabNet$ model on original feature space~(i.e., 100\% of the dimensions).} 
  \label{fig:acc_vs_dim}
\end{figure*}

We report the result of each model in \cref{fig:acc_vs_dim}  w.r.t. increasing latent dimensions: when the projection dimension increased, the accuracy of each model also increases and the inter-model difference reduces until a certain point where each model's accuracy tends to decrease again. Since embedding smaller dimensional datasets into much low dimensions tend to lose essential information to correctly classify data points, significant performance degradation is observed, e.g., in the case of health advice and forest cover type datasets, the accuracy improves up to 45 $\approx$ 55\% projected dimension only. 

In the case of very high-dimensional datasets~(e.g., GE and UJIndoorLoc), projecting them into embedding space with 5 $\approx$ 7\% dimension not to lose much information necessary to classify the data points. It is likely that higher dimensions bring more noise than information, making the classification task harder and no better than baseline~(beyond 5\% for GE and 9\% for UJIndoorLoc). 

\begin{table*}
            \centering
            \caption{Performance of individual models}
            \label{table:performance_all_models}
            \vspace{-2mm}
            \begin{tabular}{|c|l|l|l|l|l|}
            \hline
            \textbf{Classifier} & \textbf{Dataset} & \textbf{Precision} & \textbf{Recall} & \textbf{F1} & \textbf{MCC}\\ \hline
             \multirow{4}{*}{$MLP_{PCA}$}
             & Gene expression & 0.7745  &    0.7637  &  0.7553 &    0.7367\\
             & UJIndoorLoc & 0.8652  &  0.8543  &  0.8437 &    0.7741\\
             & Health advice & 0.8743  &  0.8679  &  0.8564 &    0.8067\\
             & Forest cover type & 0.7654    &  0.7547   &   0.7522  &    0.7126 \\
             \hline
             \multirow{4}{*}{$XGBoost$}
             & Gene expression & 0.8725  &    0.8623  &  0.8532 &    0.7851\\
             & UJIndoorLoc & 0.8964  &  0.8931  &  0.8836 &    0.7959\\
             & Health advice & 0.9354  &    0.9301  &  0.9155 &    0.8211\\
             & Forest cover type & 0.8382    &  0.8265   &   0.8184  &    0.7963\\
             \hline
            \multirow{4}{*}{TabNet} 
            & Gene expression & 0.9326  &    0.9276  &  0.9175 &    0.8221\\
             & UJIndoorLoc & 0.9217  &  0.9105  &  0.9072 &    0.8051 \\
             & Health advice & 0.9455  &    0.9317  &  0.9178 &    0.8235  \\
             & Forest cover type & 0.8953    &  0.8879   &   0.8854  &    0.8057 \\ 
             \hline
            \multirow{4}{*}{$SAN_{CAE}$} 
             & Gene expression & 0.9525  &    0.9442  &  0.9325 &    0.8353\\
             & UJIndoorLoc & 0.9357  &  0.9269  &  0.9175 &    0.8233 \\
             & Health advice & 0.9623  &    0.9538  &  0.9329 &    0.8451 \\
             & Forest cover type & 0.9112    &  0.9105   &   0.9023  &    0.8124 \\ 
             \hline
            \end{tabular}
\end{table*}

\begin{table*}
\centering
    \caption{Fidelity vs. confidence of rule sets for the surrogate models}
    \label{table:rules_overall_result}
    \vspace{-2mm}
    \begin{tabular}{|l|l|l|l|l|l|l|} 
        \hline
         & \multicolumn{2}{c}{\textbf{DT}} & \multicolumn{2}{c}{\textbf{RF}} & \multicolumn{2}{c|}{\textbf{ERT}}\\ 
        \hline
        \textbf{Dataset} & {Fidelity} & {Confidence} & {Fidelity} & {Confidence} & {Fidelity} & {Confidence}\\ 
        \cline{1-3}\cline{4-5}
        UJIndoorLoc & 86.16 $\pm$ 1.72 & 85.37 $\pm$ 1.53 & 89.27 $\pm$ 1.46 & 88.11 $\pm$ 1.21 & 90.25 $\pm$ 1.38 & \textbf{89.15} $\pm$ \textbf{1.57}\\ 
        \hline
        Health advice & 88.35 $\pm$ 1.45 & 87.55 $\pm$ 1.85 & 90.11 $\pm$ 1.81 & 89.45 $\pm$ 1.35 & 91.38 $\pm$ 1.65 & \textbf{90.25 $\pm$ 1.42}\\
        \hline
        Forest cover type & 90.23 $\pm$ 1.37 & 88.75 $\pm$ 1.32 & 93.15 $\pm$ 1.42 & 91.25 $\pm$ 1.31 & 94.36 $\pm$ 1.35 & \textbf{92.17 $\pm$ 1.34}\\
        \hline
        Gene expression  & 91.27 $\pm$ 1.42 & 89.33 $\pm$ 1.25 & 92.25 $\pm$ 1.35 & 90.21 $\pm$ 1.29 & 93.45 $\pm$ 1.25 & \textbf{91.37 $\pm$ 1.35}\\
        \hline
    \end{tabular}
    \end{table*}

As shown in \cref{table:performance_all_models}, the $MLP_{PCA}$ model asymptotically and consistently yielded the lowest accuracy across all datasets, while $XGBoost_{Isomap}$ slightly outperformed $MLP_{PCA}$. As PCA features are projected onto an orthogonal basis, they are linearly uncorrelated and similar to a single-layered AE with a linear activation function. Even though Isomap can learn a projection that preserve intrinsic data structure, it is not able to grasp non-linearity, hence failing to learn complex transformations. $SAN_{CAE}$ and $TabNet$ yielded comparable accuracy as both architectures learn projections that preserves relevant information for the classification. However, $SAN_{CAE}$ outperformed $TabNet$ as CAE modeled non-linear interactions among a large number of features and generate classification-friendly representations. 


\begin{figure*}[h]
  \subfloat[]{
    \includegraphics[clip,width=0.4\textwidth]{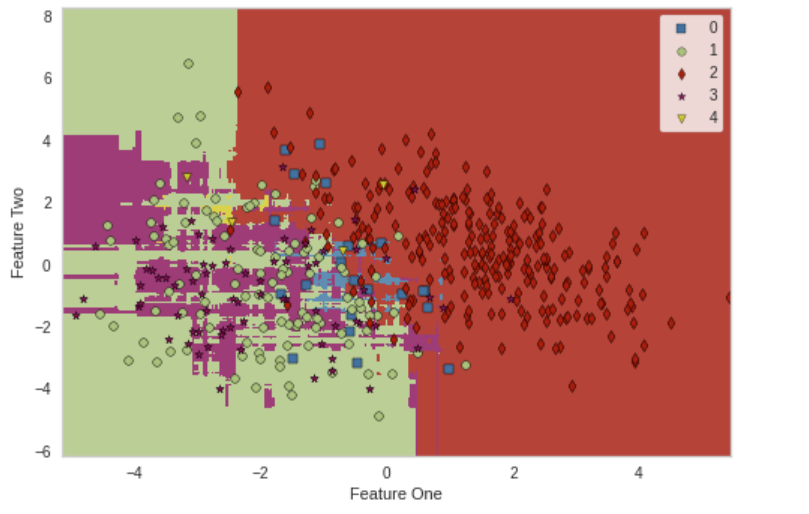}%
      \label{fig:db_DT}
  }\hfill
  \subfloat[]{
    \includegraphics[clip,width=0.36\textwidth]{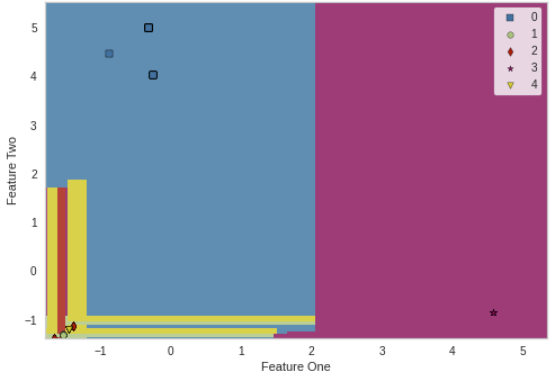}%
      \label{fig:db_RF}
  }\hfill
    \subfloat[]{
    \includegraphics[clip,width=0.36\textwidth]{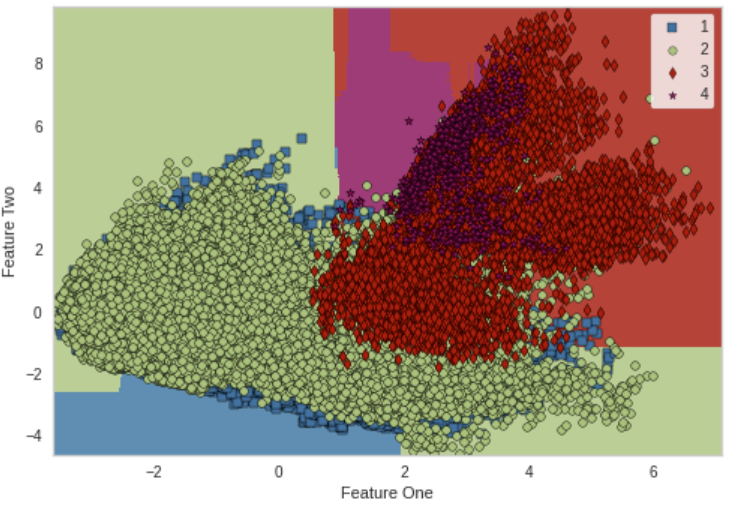}%
    \label{fig:db_GBT}
  }\hfill
  \subfloat[]{
    \includegraphics[clip,width=0.38\textwidth]{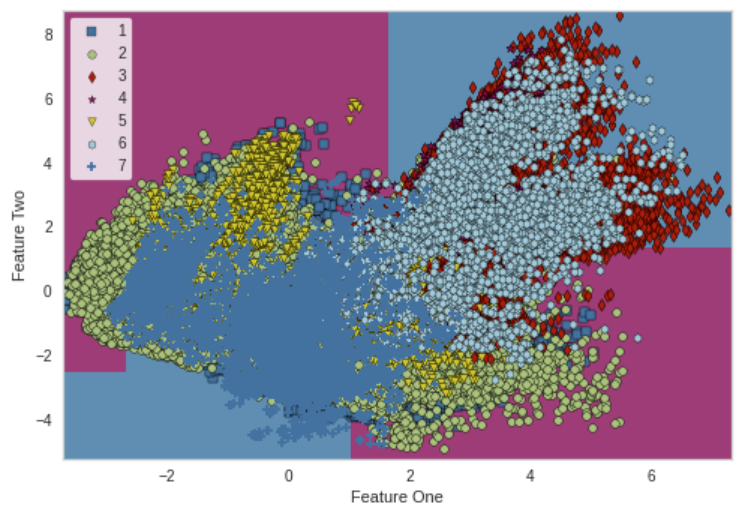}%
      \label{fig:db_ensemble}
  }
  \caption{Decision boundaries for ERT model across datasets for top-2 features: a) gene expression, b) UJIndoorLoc, c) Health advice, d) Forest cover type.} 
  \label{fig:dbs_all_models}
\end{figure*}

\begin{figure*}[h]
  \subfloat[]{
    \includegraphics[clip,width=0.23\textwidth]{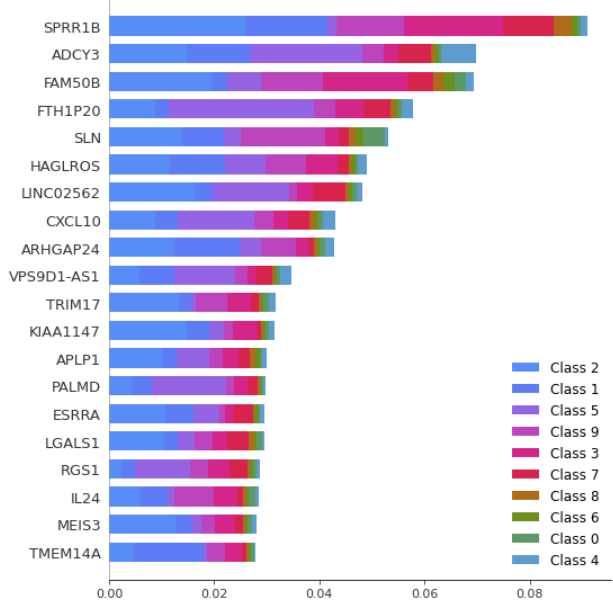}%
      \label{fig:fi_ge}
  }
  \subfloat[]{
    \includegraphics[clip,width=0.23\textwidth]{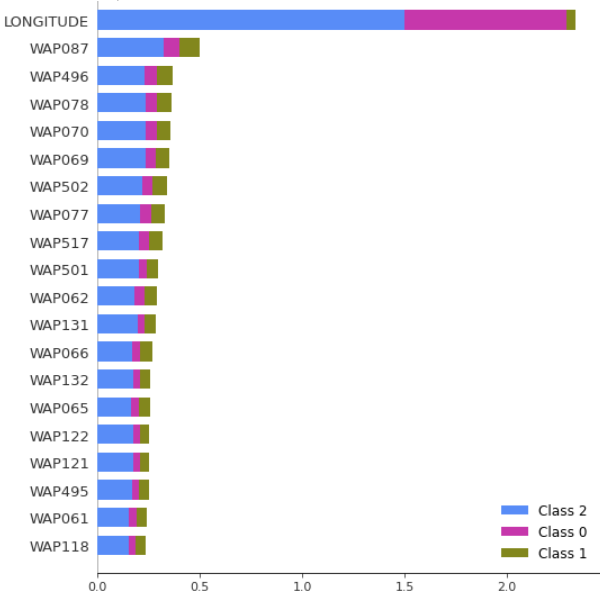}%
      \label{fig:fi_indoor}
  }
    \subfloat[]{
    \includegraphics[clip,width=0.25\textwidth]{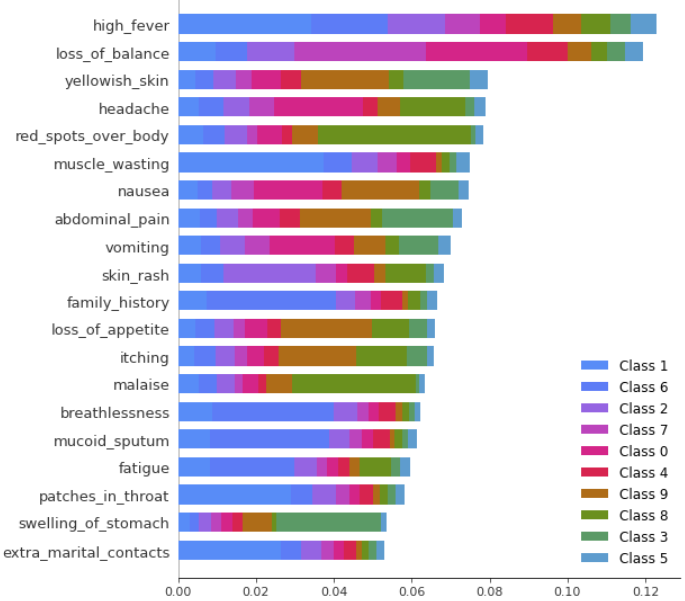}%
    \label{fig:fi_health}
  }
  \subfloat[]{
    \includegraphics[clip,width=0.27\textwidth]{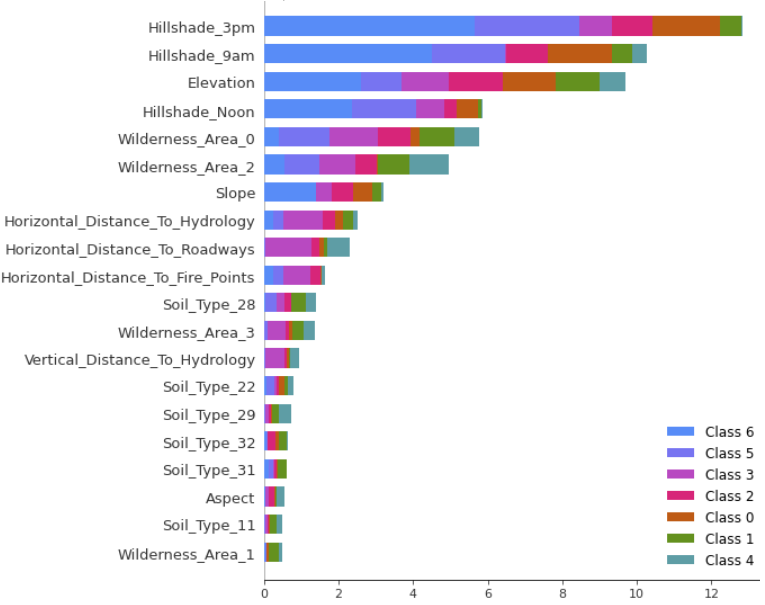}%
      \label{fig:fi_ct}
  }
  \caption{Common features per class sorted w.r.t global feature importance: a) gene expression, b) UJIndoorLoc, c) Health advice, d) Forest cover type.} 
  \label{fig:global_feature_importance}
\end{figure*}


\begin{figure*}
  \subfloat[]{
    \includegraphics[clip,width=0.45\textwidth]{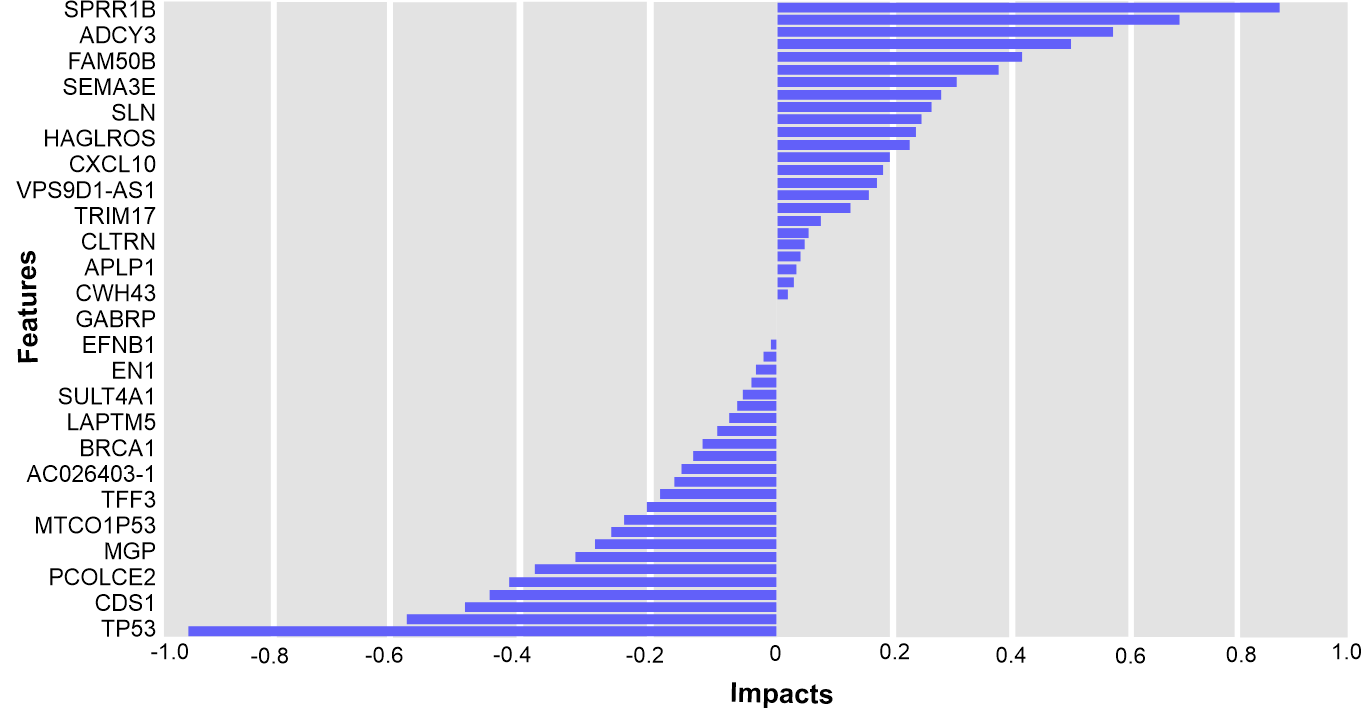}%
      \label{fig:global_feature_impacts_ge}
  }\hfil
  \subfloat[]{
    \includegraphics[clip,width=0.43\textwidth]{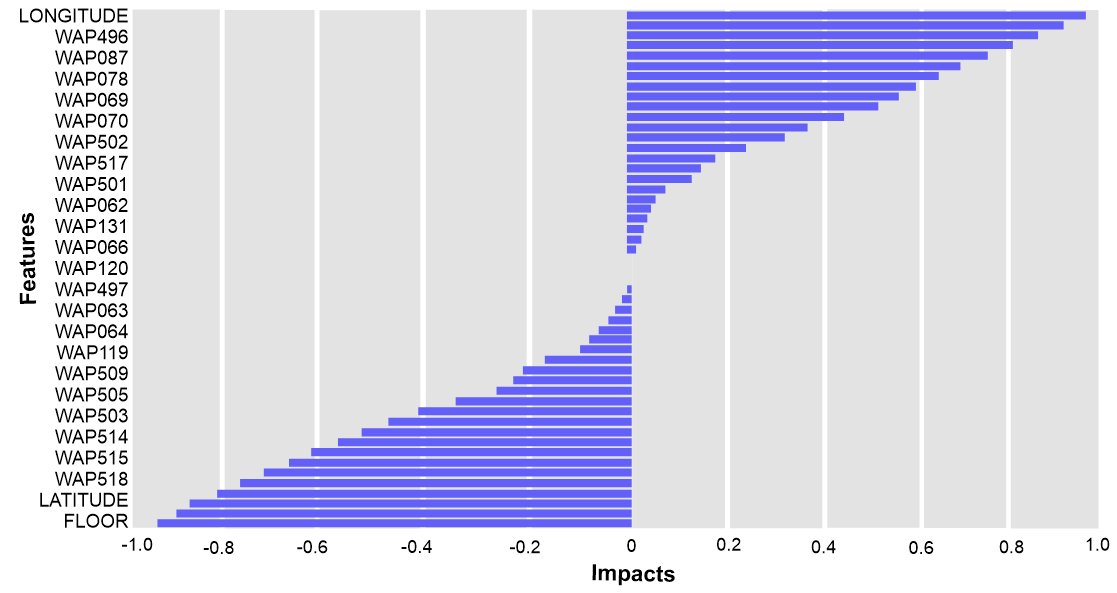}%
      \label{fig:global_feature_impacts_il}
  }\hfil
    \subfloat[]{
    \includegraphics[clip,width=0.45\textwidth]{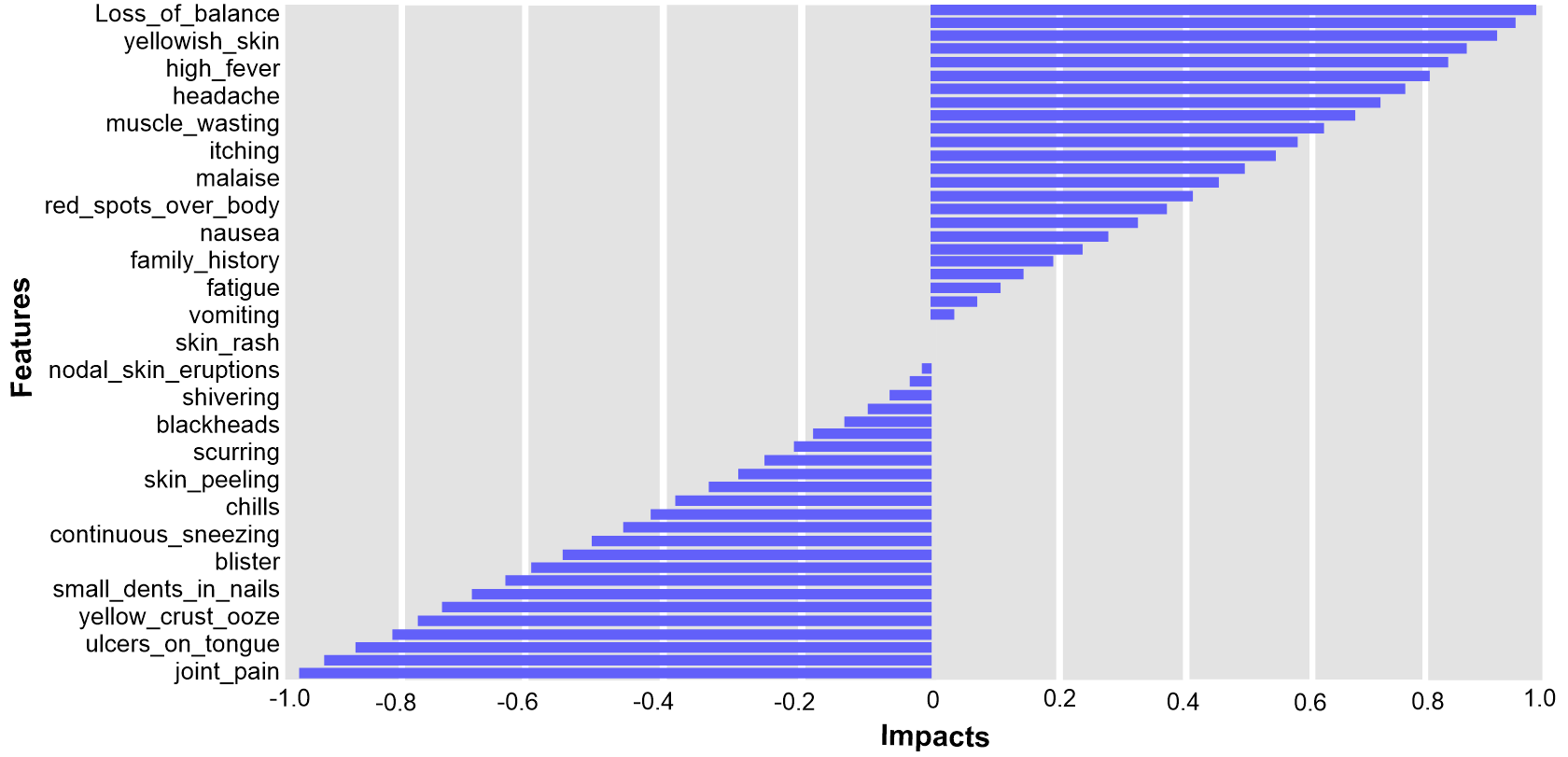}%
    \label{fig:global_feature_impacts_ha}
  }\hfil
  \subfloat[]{
    \includegraphics[clip,width=0.49\textwidth]{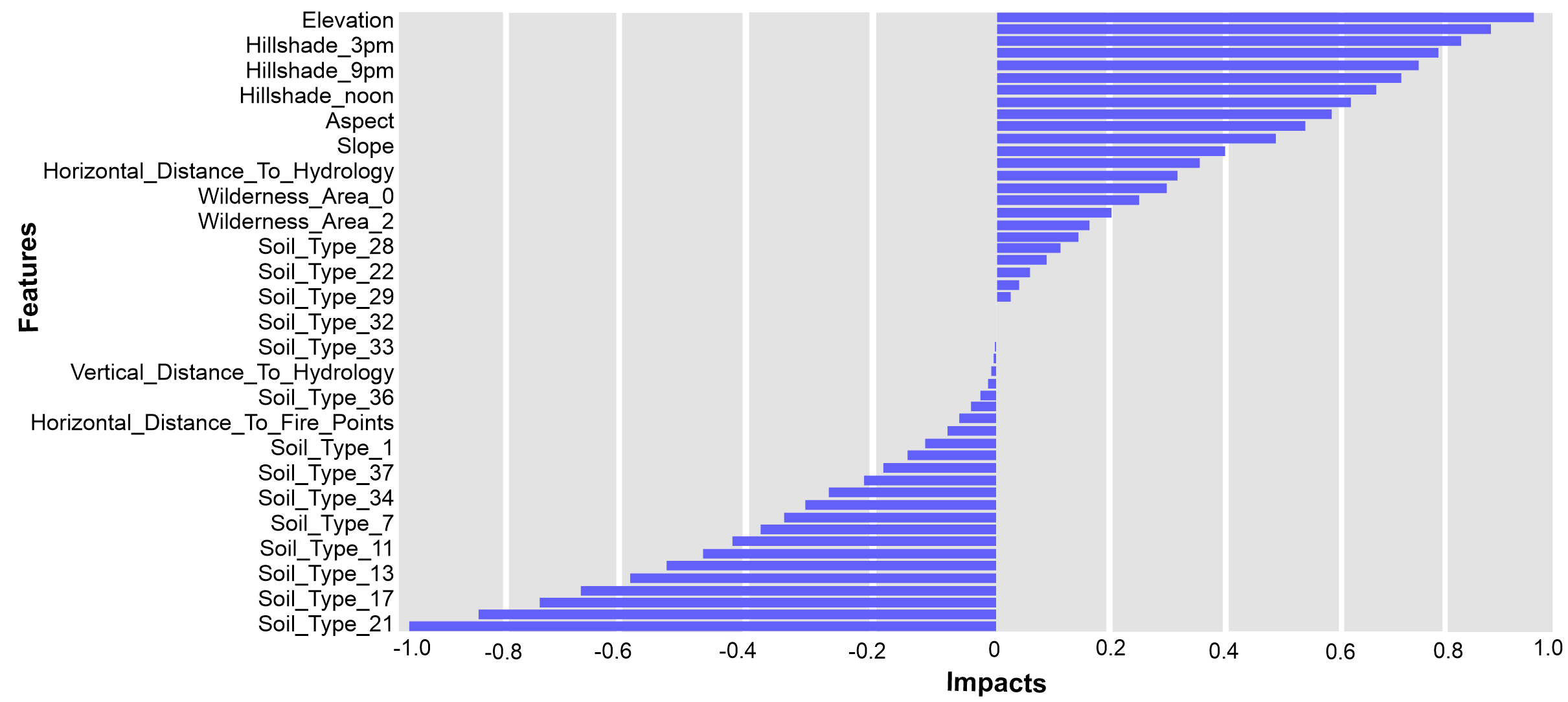}%
      \label{fig:global_feature_impacts_fct}
  }
  \caption{Global feature impacts for ERT model sorted w.r.t SHAP value: a) gene expression, b) UJIndoorLoc, c) Health advice, d) Forest cover type.} 
  \label{fig:global_feature_impacts}
\end{figure*}

The precision plot outlines the relationship between predicted probability~(that an index belongs to the positive) and the percentage of the observed index in the positive class. The observations get binned together in groups of roughly equal predicted probabilities, and the percentage of positives is calculated for each bin. Further, we provided the decision boundary~(i.e., a hyper-surface that partitions underlying vector space) by partitioning underlying vector space~(given that showing so many different classes, e.g., 33 classes for GE and 42 for the symptoms precaution datasets in a single decision space is difficult) in \cref{fig:dbs_all_models}. Each classifier classifies data points on one side of the decision boundary as belonging to one class and all those on the other side as belonging to other class. 

\begin{table*}
    \centering
    \caption{Mean variance of surrogate models}
    \label{table:r_square_}
    \vspace{-2mm}
    \begin{tabular}{|l||l||l||l|} 
        \hline
        \textbf{Dataset} & {$R^2$(\textbf{DT})} & {$R^2$(\textbf{RF})} & {$R^2$(\textbf{ERT})}\\ 
        \hline
        UJIndoorLoc & 86.16 $\pm$ 1.72 & 89.27 $\pm$ 1.46 & \textbf{91.35 $\pm$ 1.55}\\ 
        \hline
        Symptom precaution & 89.35 $\pm$ 1.45 & 92.11 $\pm$ 1.8 & \textbf{94.15 $\pm$ 1.75} \\
        \hline
        Forest cover type & 90.23 $\pm$ 1.37 & 91.15 $\pm$ 1.4 & \textbf{94.35 $\pm$ 1.25}\\
        \hline
        Gene expression & 88.25 $\pm$ 1.42 & 90.23 $\pm$ 1.35 & \textbf{93.27 $\pm$ 1.55}\\ 
        \hline
    \end{tabular}
    \vspace{-4mm}
\end{table*}


\subsection{Performance of surrogate models}
The fidelity and confidence of rule sets on test sets are demonstrated in \cref{table:rules_overall_result}. The mean fidelity is shown in percentage and the standard deviations~(SDs) for 5 runs are provided as $\pm$. Fidelity levels of 80\%, 60\% to 80\%, and below 60\% are considered as high, medium, and low, respectively. 

As of \emph{UJIndoorLoc}, the ERT model achieved the highest fidelity and confidence scores of 90.25\% and 89.15\%, with the SDs of 1.38\% and 1.57\%. RF model also performed moderately well giving the second highest scores of 88.11\% and 90.25\%, with SDs of 1.21\% and 1.38\%. 
As of \emph{health advice}, the ERT model achieved the highest fidelity and confidence of 91.38\% and 90.25\%, with the SDs of 1.65\% and 1.42\%, respectively. The RF model also performed moderately well giving the second highest fidelity and confidence of 90.11\% and 89.45\%, with slightly lower SDs of 1.81\% and 1.35\%. 

As of \emph{forest cover type}, the ERT model achieved the highest fidelity and confidence of 94.36\% and 92.17\%, with the SDs of 1.35\% and 1.34\%. The RF model also performed moderately well giving the second highest fidelity and confidence of 93.15\% and 91.25\%, with slightly lower SDs of 1.42\% and 1.31\%. As of \emph{gene expression}, the ERT model achieved the highest fidelity and confidence of 93.45\% and 91.37\%, with the SDs of 1.25\% and 1.35\%. The RF model also performed moderately well giving second highest scores of 92.25\% and 90.21\%, with slightly lower SDs of 1.35\% and 1.29\%. 

The percentage of variance~($R^2$) for the surrogate models are shown in \cref{table:r_square_}. It turns out that $R^2$ for the ERT model is comparable to the best performing $SAN_{CAE}$ and $TabNet$ models. A surrogate model that largely approximates the predictive power of a complex black-box model can be reused with some sacrifice in accuracies. 



\begin{figure*}[!h]
  \subfloat[]{
    \includegraphics[clip,width=0.45\textwidth]{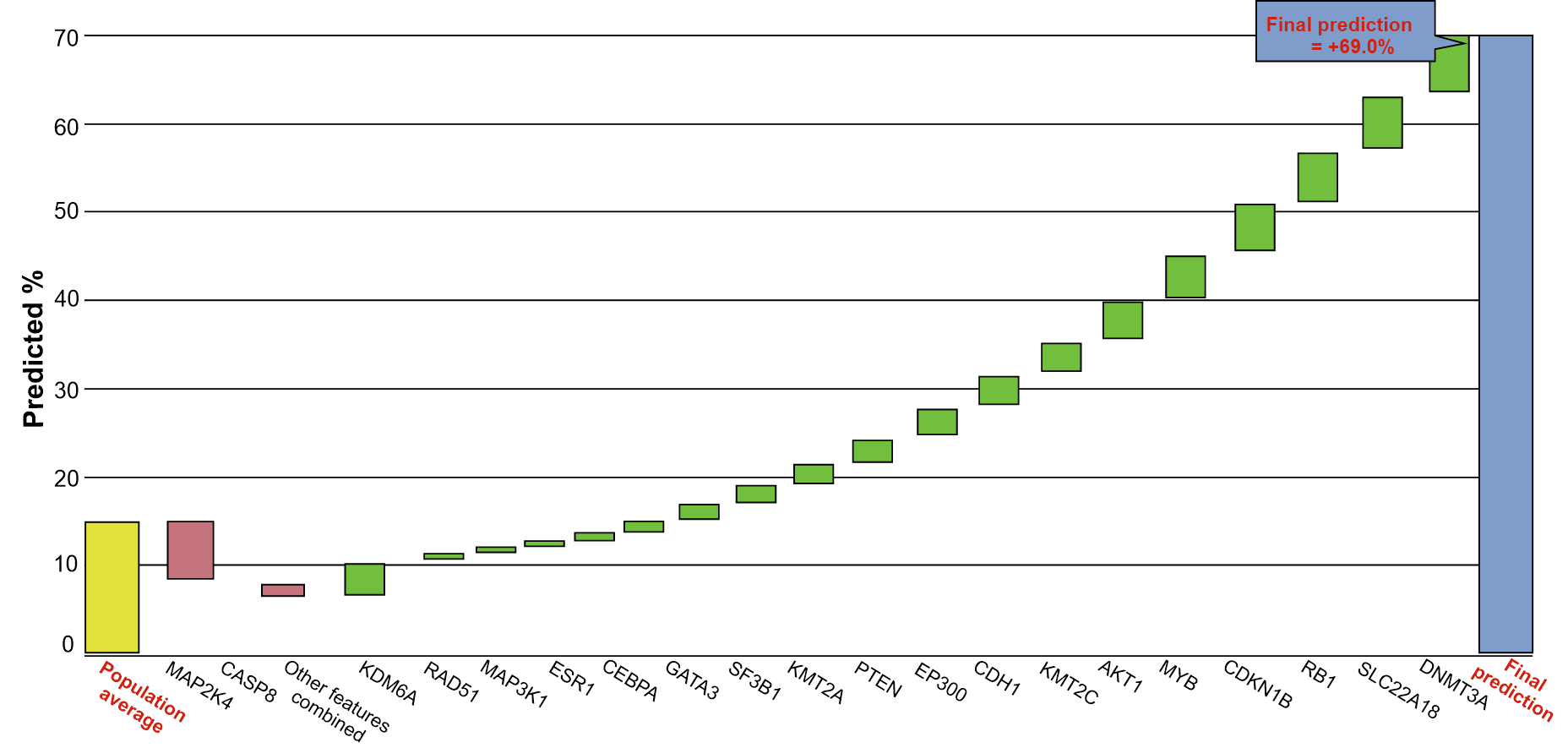}%
      \label{fig:contribution_plot}
  }\hfil
  \subfloat[]{
    \includegraphics[clip,width=0.45\textwidth]{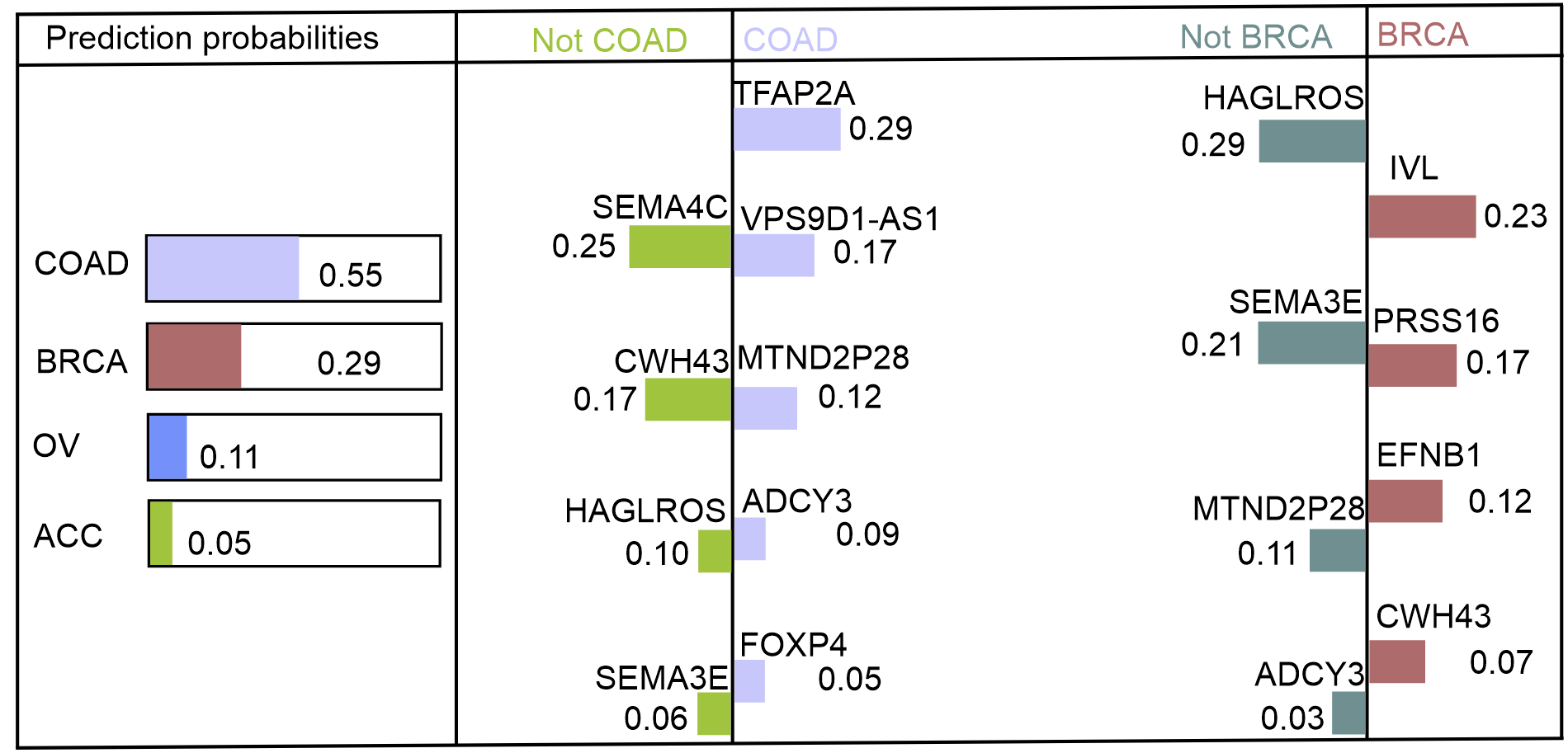}%
      \label{fig:what_if_leave_one_feature_out}
  }
  \caption{Validation of top-k feature for a specific observation: a) individual feature contribution, b) feature-level relevance.} 
  \label{fig:xxx}
\end{figure*}

\subsection{Global interpretability}
Providing both global and local explanations for all the datasets will be overwhelming. Therefore, we focus on the gene expression dataset only. Accurate identification of the most and least significant features help understand various aspects such as their relevance w.r.t certain predictions or classes. For example, biologically relevant genes w.r.t specific cancer types provide insights into carcinogenesis as they could be viewed as potential biomarkers. Therefore, both GFI and impacts are analysed to understand model's behavior. Common and important features~(w.r.t GFI score) identified with the $SAN_{CAE}$ model are demonstrated in \cref{fig:global_feature_importance}, where GFI assign a score to input features based on how useful they are at predicting a target class or overall classes. 

However, unlike feature impact, feature importance does not provide which features in a dataset have the greatest positive or negative effect on the outcomes of a model. Therefore, global feature impacts for the $SAN_{CAE}$ model sorted from most to least important w.r.t SHAP value are shown in \cref{fig:global_feature_impacts}. As seen, SHAP gives slightly different views on individual feature impacts: SPRR1B, ADCY3, FAM50B, SEMA3E, SLN, HAGLROS, CXCL10, VPS9D1-AS1, TRIM17, CLTRN, APLP1, and CWH43 positively impact the prediction, which signifies that if the prediction is in favor of specific cancer type~(e.g., \emph{COAD}) for a specific patient, these variables will play a crucial role in maintaining this prediction. On the other hand, TP53, CDS1, PCOLCE2, MGP, MTCO1P53, TFF3, AC026403-1, BRCA1, LAPTM5, SULT4A1, EN1, EFNB1, and GABRP have negative impacts on the prediction, which means if the prediction is \emph{COAD} and the value of these variables are increased, the final prediction is likely to end up flipping to another cancer type.

\subsection{Local interpretability}
First, we randomly pick a sample from the test set and perform the inferencing with the ERT model. Assuming the model predicts the chosen instance is of \emph{COAD} cancer type, the contribution plot shown in \cref{fig:contribution_plot} outlines how much contribution each feature had for this prediction. Features~(genes) DNMT3A, SLC22A18, RB1, CDKN18, MYB are top-k features according to impact values, while features CASP8 and MAP2K4 had negative contributions. 
Further, to quantitatively validate the impact of top-k features and to assess feature-level relevances, we carry out the \emph{what-if} analysis. 

\begin{figure*}[!h]
	\centering
	\includegraphics[width=0.6\textwidth]{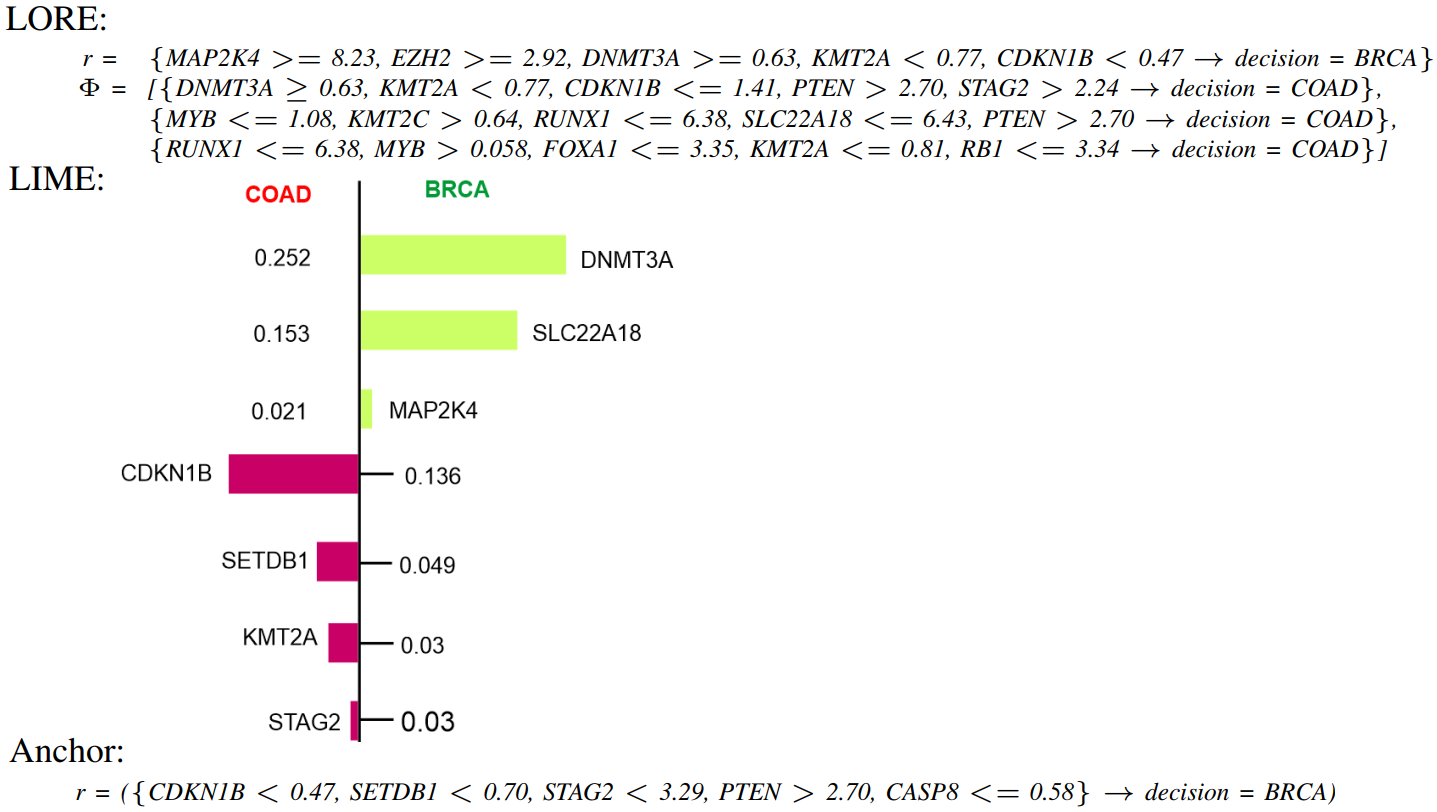}
	\caption{Example of explaining single prediction using rules, counterfctuals, and additive feature attributions.} 
	\label{fig:expl_example}
\end{figure*}

As shown in \cref{fig:what_if_leave_one_feature_out}, the observation is of \emph{COAD} with a probability of 55\% and \emph{BRCA} type with a probability of 29\%. Features on the right side~(i.e., TFAP2A, VPS9D1-AS1, MTND2P28, ADCY3, and FOXP4 are positive for \emph{COAD} class, where feature TFAP2A has the highest positive impact of 0.29) positively impact the prediction, while features on the left negatively. Genes TFAP2A, VPS9D1-AS1, MTND2P28, ADCY3, FOXP4, GPRIN1, EFNB1, FABP4, MGP, AC020916-1, CDC7, CHADL, RPL10P6, OASL, and PRSS16 are most sensitive to making changes, while features SEMA4C, CWH43, HAGLROS, SEMA3E, and IVL are less sensitive to making changes. However, if we remove feature TFAP2A from the profile, we would expect the model to predict the observation to be of \emph{COAD} cancer type with a probability of 26\%~(i.e., 55\% $-$ 29\%). This will recourse the actual prediction to \emph{BRCA} in which features IVL, PRSS16, EFNB1, and CWH43 are most important, having impacts of 0.23, 0.17, 0.123, and 0.07, respectively. These features not only reveal their relevance for this decision but also signify that removing them is like to impact the final prediction. 

Further, we focus on local explanations for this prediction by connecting \emph{decision rules} and counterfactuals with additive feature attributions~(AFA) in \cref{fig:expl_example}. While Anchor provides a single rule outlining which features impacted at arriving this decision, LIME generates AFA stating which features had positive and negative impacts. However, using decision rules and a set of counterfactuals, we show how the classifier could arrive at the same decision in multiple ways due to different negative or positive feature impacts.

\section{Conclusion}\label{sec:con}
In this paper, we proposed an efficient technique to improve the interpretability of complex black-box models trained on high-dimensional datasets. 
We found that embedding very high-dimensional datasets~(e.g., gene expression and UJIndoorLoc) into 5 $\approx$ 7\% embedding space helps improve classification accuracy as it is less likely to lose essential information to correctly classify the data points. However, higher dimensionality tends to bring more noise than information, complicating the classification task. in the case of comparatively smaller dimensional datasets~(e.g., health advice and forest cover type), embedding them into fewer dimensions tend to lose essential information, resulting in performance degradation. For example, having it projected into 45 $\approx$ 55\% embedding dimensions, the model accuracy tends to improve. 

Our model surrogation strategy is equivalent to the knowledge distillation process for creating a simpler model. However, instead of training the \emph{student} model on \emph{teacher's} predictions, we transferred learned knowledge ~(e.g., top-k or globally most and least important features) to a student and optimize an objective function. 
Further, the more trainable parameters are in a black-box model, the bigger the size of the model would be. This makes the deployment infeasible for such a large model in resource-constrained devices\footnote{For example, IoT devices having limited memory and low computing power.}. Further, the inferencing time of such a large model increases and ends up with poor response times due to network latency even when deployed in a cloud computing infrastructure, which is unacceptable in many real-time applications. 

We hope that our model surrogation strategy would help create simpler and lighter models and improve interpretability in such a situation. 
Nevertheless, we would argue that depending on the complexity of the modeling tasks, a surrogate model may not be able to fully capture the behavior of complex black-box models. Consequently, it may lead users to wrong conclusions -- especially if the knowledge distillation process is not properly evaluated and validated. In the future, we want to focus on other model compression techniques such as quantization~(i.e., reducing numerical precision of model parameters or weights) and pruning~(e.g., removing less important parameters or weights w.r.t overall accuracy). 

\bibliographystyle{splncs04}
\bibliography{references.bib}

\section*{Ethical implications}
All the data used for this study were collected from openly available sources. Besides, authors would like to clarify that no manually crafted features (content moderation or adversarially created) or data other than mentioned sources~(e.g., external datasets) we included that could potentially introduce any bias. This study did not include or process any personal or privacy-sensitive information. The scope is strictly scientific and and not intended to hurt or offense any specific person, entity, or religious/political groups/parties. Besides, authors would like to clarify that the potential use of this study is not intend for policing or the military, but just come up with a proof-of-study.

\end{document}